# Distributed Formation Control for Autonomous Robots in Dynamic Environments

Anh Duc Dang, Institute of Control Engineering, University of the Federal Armed Forces Hamburg , Holstenhofweg 85, D-22043 Hamburg, Germany
Hung Manh La[*], Department of Computer Science and Engineering, University of Nevada, Reno, NV 89557, USA
Thang Nguyen, Department of Electrical Engineering and Computer Science, Cleveland State University, Cleveland, OH 44115, USA
Joachim Horn, Institute of Control Engineering, University of the Federal Armed Forces Hamburg , Holstenhofweg 85, D-22043 Hamburg, Germany

In this paper, we propose a novel and distributed formation control method for autonomous robots to follow the desired formation while tracking a moving target in dynamic environments. In our approach, the desired formations, which include the virtual nodes arranged into specific shapes, are first generated. Then, autonomous robots are controlled by the proposed artificial force fields in order to converge to these virtual nodes without collisions. The stability analysis based on the Lyapunov approach is given. Moreover, a new combination of rotational force field and repulsive force field in designing an obstacle avoidance controller allows the robot to avoid and escape the convex and nonconvex obstacle shapes. The V-shape and circular shape formations with their advantages are utilized to test the effectiveness of the proposed method.

**ACM Reference format:**

## 1 INTRODUCTION

In recent years, multi-agent systems have widely been researched in many areas, such as physic, biology, cybernetics, and automatic control over the world. Formation control is one of the necessary and important problems in the research field on multi-agent systems. The formation control of autonomous robots, such as underwater vehicles [10], unmanned aerial vehicles [11], [54], mobile sensor networks [12]-[23], rectangular agents [50], nonholonomic mobile robots [49], [51]- [53], has potential applications in search and rescue missions, forest fire detection and surveillance, etc.

Centralized control protocols have been constructed based on the common assumption that the information of all agents is available or the multiagent system possesses all-to-all communication. The drawbacks of the centralized communication control architect are inflexibility and large computational costs for each controller for each agent especially when the number of robots is large. In contrast, a distributed control approach can provide more flexibility, easier implementation, and less computation loads as the controller of each agent only requires the information of its neighbor agents [15, 18, 39, 40, 41, 46].

In formation control of a multiagent system, the main aim is that a robot team must work together in order to achieve desired tasks, such as tracking and observing a moving target. Formation control of autonomous robots is inspired from natural behavior of fish schooling, bird flocking or ant swarming, and guarantees that the members in the formation have to move together to satisfy some conditions such as velocity matching and collision avoidance. There are several methods to generate and control the formation of a swarm of mobile robots. The artificial potential field method is known as a positive tool in order to control the coordination and the motion of a swarm towards the target position, see [1]-[9], [13]. It is well established that the potential field generates impulsive/attractive forces for mobile robots to avoid collision and maintain distances in coordination control problems [13]. The success of the formation control method based on the random connections among neighboring members in a swarm as an α-lattice configuration has been published in some literature, such as [12]-[25]. In this method, the neighboring robots are linked to each other by the attractive/repulsive force fields among them to create a robust formation without collisions. On the other hand, the formation control based on dynamic framework was introduced [28, 29] in which the robots can adjust their formation by rotating and scaling during their movement. In another approach, robots are controlled to achieve given positions in the desired shape [30, 31]. Although the artificial potential field is known as a positive method for path planning of mobile robots, this approach possesses some limitations due to local minimum problems. Namely, when the attractive force of the target and the repulsive force of the obstacles are equal and collinear but in an opposite

TABLE I

SUMMARY OF RELATED WORK IN MULTI-AGENT FORMATION CONTROL

| α-lattice connection formation (LCF) | Dynamic framework formation (DFF) | Distributed formation following desired shapes (DFS) |
|---|---|---|
| Flocking of mobile sensor networks [12]-[16], flocking of multi agent systems [17]-[21], and adaptation and stability of a swarm in a dynamic environment [22]-[25]. | Formation control following a framework [26], [27], [30], [31], and formation control following dynamic region [28], [29]. | Our paper presents a new approach to multi-robot formation control following a desired shape formation (V-shape [32]-[34] or circular shape [35], [36], etc.) to track, and encircle a moving target under the effect of the dynamic environment. |
| LCF is mainly designed based on the random connections among the neighboring members in an α-lattice configuration. The convergence and adaptation of a swarm in a dynamic environment were verified. How-ever, when we need a practical formation, such as V-shape, circular shape or linear formation, etc., LCF method is still constrained. In this method, each robot's motion depends on the motion of its neighbors. | In the DFF method, all robots in the group move together within a given framework or region. They stay within a moving region and are able to adjust their formation by rotating and scaling during their movement. This method does not require specific orders or positions of all robots inside the given region. Each robot's motion depends on the motion of its neighbors and framework. | In our approach, the shape of the desired formation is easily designed. Robots are independently controlled by the attractive potential field from the virtual nodes in the desired formation. Moreover, the designed virtual nodes guarantee that the neighboring robots do not interact to each other. Hence, robots easily converge to these virtual nodes under the velocity matching. Using an added rotational vector field, robots can quickly escape convex and concave obstacles to continue to track the target. |

direction, the total force on the robot is equal to zero. Hence, this causes the robot motion to stop. Moreover, in complex environments with convex and concave obstacle shapes, such as U-shaped obstacles or long walls, etc., the application of the traditional potential field method is conservative. Robots can be trapped in these obstacles before reaching the target, see [8], [9]. The summary of related work is shown in Table I. Recent research methods were proposed to address local minima avoidance such as [44], [45] that the reader can refer to.

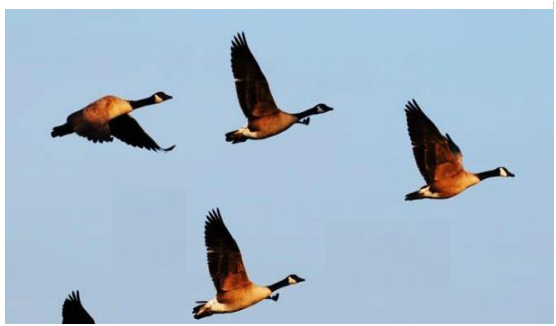

Fig. 1. The V-shape flying formation of birds.
(Source: http://www.grahamowengallery.com/photography/geese.html)

The main contributions of this paper are as follows. The distributed formation control algorithms are designed to drive multiple robots to converge to the desired positions in dynamic environments. These control algorithms guarantee that the stability of the formation is maintained, and there are no collisions among robots while tracking a moving target. Furthermore, the obstacle avoidance control algorithm is built based on the combination of the rotational force field and the repulsive force field surrounding the obstacles to drive robots to escape these obstacles without collisions. The stability analysis of the proposed control algorithms based on Lyapunov approach is given.

The remaining sections of this paper are organized as follows. The problem formulation is presented in Section 2. In Section 3, the formation control algorithms are presented. Section 4 presents a case study for two typical examples of V-shape and circular shape formation control. Simulation results are discussed in Section 5. Finally, Section 6 concludes the paper and proposes future research topics.

## 2 PROBLEM FORMULATION

We consider a swarm of $N$ robots and their mission to track a moving target in two-dimensional space. Let $p_i = (x_i, y_i)^T$, $v_i = (v_{ix}, v_{iy})^T$ be the position, velocity vectors of the robot $i$ ($i=1,2,3,...,N$), respectively. The dynamic model of the robot $i$ is described as

$$\dot{p}_i = v_i$$
$$\dot{v}_i = u_i, \quad i = 1,..,N. \tag{1}$$

A formation of autonomous robots must satisfy the following conditions:

a. There are no collisions among robots.
b. Robots must converge to the desired positions.
c. Robots must avoid collision with obstacles.
d. The formation must be maintained under the influence of the environment.

A desired formation of a multi-robot system consists of virtual nodes which will be occupied by the robots of the system.

To facilitate the control design and analysis, we introduce the following definitions:

**Definition 1.** Robot $i$ ($i=1,2,3,...,N$) is called an active robot at time $t$ if the distance from it to virtual node $j$ ($j=1,2,3,...,N$) is smaller than the radius of the active circle surrounding each virtual node ($d_i^j < r_a$, $r_a = d/2 - \lambda_a$, $\lambda_a$ is a positive constant), see Fig. 2. Otherwise, it is a free robot.

**Definition 2.** Virtual node $j$ ($j=1,2,..,N$; $q_j=(x_j, y_j)^T$; $v_j=(v_{jx}, v_{jy})^T$) of the desired formation is active if there is a robot $i$ ($i=1,2,..,N$) in the active circle of this virtual node, see Fig.2. In contrast, it is free.

**Definition 3.** A desired position for each robot $i$ in the desired formation is a virtual node $j$ at which $\lim_{t \to \infty} (\|p_i(t) - q_j(t)\|) = 0$, and the virtual node ($j$-$1$) is also active.

**Definition 4.** The desired V-shape formation is a formation that is linked by two line formations. These line formations are driven by a selected leader, and connected by a formation angle $\varphi$. In each line formation, the virtual nodes are equidistant to each other.

**Definition 5.** The desired circular shape formation is the locus of all virtual nodes that are equidistant to each other and equidistant from the target.

**Definition 6.** A noisy environment is the environment in which the measurement of agents is affected by some noise.

**Definition 7.** Let $N_i^k(t)$ be the set of the neighboring robots of the robot $i$ at time $t$, such that:

$$N_i^k(t) = \{\forall k : d_i^k = \|p_i - p_k\| \leq r_r, k \in \{1,..N\}, k \neq i\}, \quad (2)$$

where $r_r$ is the repulsive radius surrounding each robot, and $d_i^k$ is the Euclidean distance between robot $k$ and robot $i$.

**Definition 8.** A set of the neighboring obstacles of robot $i$ at time $t$ is defined as

$$N_i^o(t) = \{\forall o : d_i^o \leq r^\beta, o \in \{1,..M\}, o \neq k\}, \quad (3)$$

where $r^\beta > 0$ and $d_i^o = \|p_i - p_o\|$ are the obstacle detection range and the Euclidean distance between robot $i$ and obstacle $o$, respectively.

**Definition 9.** The equilibrium point x = 0 of $\dot{x} = f(x)$ is [47]

- stable if for each $\epsilon > 0$ there is $\delta > 0$ (dependent on $\varepsilon$) such that $\|x(0)\| < \delta \Rightarrow \|x(0)\| < \varepsilon, \forall t \geq 0$.
- unstable if it is not stable.
- asymptotically stable if it is stable and $\delta$ can be chosen such that $\|x(0)\| < \delta \Rightarrow \lim_{t \to \infty} \|x(t)\| = 0$.

## 3 CONTROL ALGORITHMS

This section presents control algorithms to guarantee that all robots will converge to the desired positions in the desired formation. While tracking a moving target, the stability of the formation must be maintained, and there are no collisions among members. Additionally, robots must automatically escape the obstacles in order to continue to track the moving target with their swarm.

### 3.1 Control architecture

To address our problem, we propose the control input $u_i$ for each robot as follows:

$$u_i = \begin{cases} u_i^t + u_i^o, & \text{if robot } i \text{ is leader } l; i, l = 1,..., N \\ u_i^j + u_i^o + u_i^k, & \text{otherwise} \end{cases} \quad (4)$$

where the first controller $u_i^j$ is used to control the formation connection, the second controller $u_i^o$ drives robots to avoid obstacles, and the controller $u_i^k$ is added to help robots avoid collision during their movement. Using the tracking controller $u_i^t$, the leader can easily drive its swarm towards the target. In (4), when collision avoidance for robot $i$ is active when other robots lie in its repulsive radius $r_r$, which activates $u_i^k$. Similarly, when an obstacle lies in the obstacle sensing range $r^\beta$, obstacle avoidance is active, leading to the activation of $u_i^o$.

The stability analysis of the main results in this paper will be based on the following Lyapunov's stability theorem [38], [47].

**Theorem 1.** Let $V(x)$ be a continuously differentiable function defined in a domain $D \subset R^n; 0 \in D$. If there is $V(x)$ such that $V(0) = 0, V(x) > 0, \forall x \in D/\{0\}$, and $\dot{V}(x) \leq 0, \forall x \in D$, then the origin is a stable. Moreover, if $\dot{V}(x) < 0, \forall x \in D/\{0\}$, then the origin is asymptotically stable. Furthermore, if $V(x) > 0, x \neq 0, \|x\| \to \infty \Rightarrow V(x) \to \infty$, and $\dot{V}(x) < 0, x \neq 0$, then the origin is globally asymptotically stable. $V(x)$, which satisfies the conditions of the theorem, is called a Lyapunov function.

## 3.2 Formation connection control algorithm

Firstly, surrounding the virtual nodes $j$ ($j=1,2,...N$), the attractive force fields are created to drive free robots to move towards their desired positions. Then, these free robots will occupy these desired positions, and become active robots. The tracking task is to make the distance $d_i^j = \|p_i - q_j\|$ approach zero as fast as possible. This means that $\lim_{t\to\infty}(\|p_i(t) - q_j(t)\|) = 0$ and $\lim_{t\to\infty}(\|v_i(t) - v_j(t)\|) = 0$.

Therefore, the formation control law for formation connection is proposed in Algorithm 1.

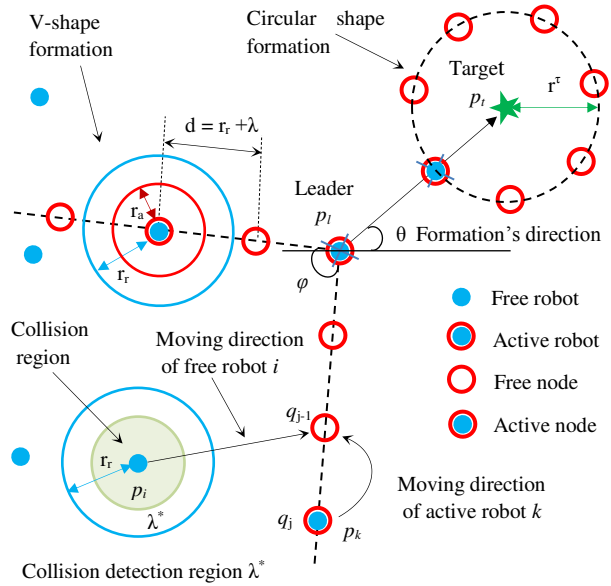

Fig. 2. Description of the formation control method following the V-shape and circular shape desired structure.

In Algorithm 1, $k_{ip1}^j, k_{ip2}^j, k_{ip3}^j, k_{iv}^j$, $(p_i - q_j)$ and $(v_i - v_j)$ are the positive constants, the relative position vector, the relative velocity vector between the robot $i$ and the virtual node $j$, respectively. In this algorithm, we use two potential fields $f_{1i}^j = -k_{ip3}^j (p_i - q_j)/\|p_i - q_j\|$ and $f_{2i}^j = -k_{ip1}^j (p_i - q_j)$ as the artificial attractive forces. The constant potential field $f_{1i}^j$ is used to drive the free robots towards the desired formation, while the linear potential field $f_{2i}^j$ is used to control the active robots to approach the virtual nodes. Additionally, the component $-k_{iv}^j (v_i - v_j)$ is utilized as the damping term. Therefore, using the Algorithm 1, robots can quickly approach the desired positions at the virtual nodes of the desired formation. The stability of the formation is stated in the following.

**Theorem 2.** *Consider the active robot i with its dynamic model (1) and control input $u_i^j$ given in Algorithm 1 at the active node j in the desired formation. If the velocity of node j is smaller than the maximum velocity of robot*

$i$, and node $j-1$ is also active, then the system (1) will converge to the equilibrium state, at which $p_i = q_j$ and $v_i = v_j$ for all $i$ and $j$.

**Proof:** See the proof of the theorem is given in Appendix.

**Remark 1.** It is well established that the Lyapunov theory is employed for the stability proof of formation and flock control of multiagent systems [13]. For more information about the Lyapunov theory, the reader is referred to [38], [47].

**Remark 2.** Fig. 2 illustrates a desired formation (V-shape or circular shape formation) of $N$ virtual nodes. Each robot must find a desired position in this desired formation. Firstly, each free robot $i$ will pursue the closest free virtual node $j$ in order to be active at this virtual node. If the position of an active robot at the active node $j$ is still not the desired position (for example robot $k$ in Fig. 2 with $k=1,2,..,N$, $k \neq l$), then this active robot will automatically move into the virtual nodes $(j-1)$ until it achieves a desired position.

In order to allow robot $i$ to approach node $j$ as fast as possible, we use the attractive force from node $j$, which is proportional to its velocity. This means that the factor $k_{ip1}^j$ in Algorithm 1 depends on the velocity $v_j$. This factor is given as

$$k_{ip1}^j = k_{ipd}^j + \varepsilon_2 \|v_j\| \tag{5}$$

where $k_{ipd}^j$, $\varepsilon_2$ are positive.

---

**Algorithm 1**: Reaching the desired positions at the virtual nodes in the desired formation

---

**Consider:** A robot $i$ and virtual nodes $j$ ($i,j=1,..,N$, $i \neq l$). Determine the shortest distance from $p_i$ to all the virtual nodes $q_j$ and the scalar constant $c_i^j$ such that:

$$d_i^{jm1} = \min\{d_i^j = \|p_i - q_j\|, j = 1,..,N\}, c_i^j = \begin{cases} 1 & \text{if } q_j \text{ is active} \\ 0 & \text{if } q_j \text{ is free}. \end{cases}$$

**if** $d_i^{jm1} \leq r_a$ & $c_i^{jm1-1} = 1$ **then**
$$u_i^j = -k_{ip1}^j (p_i - q_{jm1}) - k_{iv}^j (v_i - v_{jm1}) + \dot{v}_{jm1}$$
**else if** $d_i^{jm1} \leq r_a$ & $c_i^{jm1-1} = 0$ **then**
$$u_i^j = -k_{ip2}^j (p_i - q_{jm1-1}) - k_{iv}^j (v_i - v_{jm1-1}) + \dot{v}_{jm1-1}$$
**else if** $d_i^{jm1} > r_a$ **then**
    **if** $c_i^{jm1} = 0$ **then**
$$u_i^j = -k_{ip3}^j (p_i - q_{jm1}) / \|p_i - q_{jm1}\| - k_{iv}^j (v_i - v_{jm1}) + \dot{v}_{jm1}$$
    **else**
    Determine the shortest distance from $p_i$ to the free virtual nodes $q_j$ in the desired formation as
$$d_i^{jm2} = \min\{d_i^j = \|p_i - q_j\|, c_i^j = 0, j = 1,..,N, j \neq jm1\}$$
$$u_i^j = -k_{ip3}^j (p_i - q_{jm2}) / \|p_i - q_{jm2}\| - k_{iv}^j (v_i - v_{jm2}) + \dot{v}_{jm2}$$
    **end**
**end**

### 3.3 Collision avoidance control algorithm

This section presents a method for the collision avoidance among the robots during their movement based on the artificial repulsive potential field. The potential field has been widely used in addressing flocking and formation control of multiagent systems [1]-[9], [13], [19]. In general, a potential function $\psi(.)$, which depends on the distance among agents or between agents and obstacles, is constructed. A typical feedback control law consists of two parts: the gradient of the potential function and the velocity consensus term. For more details, see [1]-[9], [13], [19], [39] and references therein.

In order to avoid the collision between robots $i$ and $k$ ($i, k=1,2,..,N$; $i \neq k \neq l$), the local repulsive force field is created surrounding each robot within the repulsive radius $r_r$ as

$$f_i^k = \left( \left( \frac{1}{d_i^k} - \frac{1}{r_r} \right) \frac{k_{i1}^k}{(d_i^k)^2} - k_{i2}^k \left( d_i^k - r_r \right) \right) c_i^k n_i^k, \tag{6}$$

where the positive factors $k_{i1}^k$, $k_{i2}^k$ are used to control the fast interaction. The unit vector $n_i^k$ from robot $k$ to robot $i$ is given as $n_i^k = (p_i - p_k)/\|p_i - p_k\|$. The scalar $c_i^k$ is defined as follows:

$$c_i^k = \begin{cases} 1 & \text{if } k \in N_i^k(t) \\ 0 & \text{otherwise.} \end{cases} \tag{7}$$

The algorithm for the collision avoidance is built based on the repulsive vector field (6) combined with the relative velocity vector $k_{iv}^k (v_i - v_k)$ between robot $k$ and robot $i$ as follows:

$$u_i^k = \sum_{k=1, k \neq i}^{N} \left( f_i^k - c_i^k k_{iv}^k (v_i - v_k) \right). \tag{8}$$

The controller (10) shows that the neighboring robots are always driven to leave each other. In other words, this controller guarantees that there are no collisions among robots in the swarm.

### 3.4 Obstacle avoidance control algorithm

This section presents a control algorithm for robots to pass through $M$ obstacles. The obstacle avoidance control algorithm for each member robot $i$ ($i=1,2,..,N$) is designed as follows

$$u_i^o = \sum_{o=1, o \neq k}^{M} \left( f_i^{op} + f_i^{or} + k_{iv}^o c_i^o (v_i - v_o) \right) \tag{9}$$

where the relative velocity vector $(v_i - v_o)$ between robot $i$ and its neighboring obstacle $o$ is used as a damping term with damping scaling factor $k_{iv}^o$, and the scalar $c_i^o$ is defined as:

$$c_i^o = \begin{cases} 1 & \text{if } o \in N_i^o(t) \\ 0 & \text{otherwise.} \end{cases} \tag{10}$$

The repulsive force field $f_i^{op}$ is created to drive robot $i$ to move away from its neighboring obstacle, see Fig.4. This force field is designed as:

$$f_i^{op} = c_i^o \left( \left( \frac{1}{d_i^o} - \frac{1}{r^\beta} \right) \frac{k_{i1}^o}{(d_i^o)^2} - k_{i2}^o \left( d_i^o - r^\beta \right) \right) n_i^o \tag{11}$$

where the positive factors $k_{i1}^o$, $k_{i2}^o$ are used to control the fast obstacle avoidance, $r^\beta$ is the obstacle detection range defined in Definition 8, and $n_i^o = (p_i - p_o)/\|p_i - p_o\|$ is a unit vector.

In control law (9), the rotational force field $f_i^{or}$ (see Fig. 4) is added to combine with the repulsive force $f_i^{op}$ (see Fig. 4) to drive robot $i$ to quickly escape its neighboring obstacle. While the potential force field is used to enable it to avoid collision with its neighboring obstacle, the rotational force field is used to solve the local minimum problems, for instance, the robot meets a trapping point, at which the repulsive force of the obstacles and the attractive force of the target are balanced. Under the effect of the added rotational force field, the robot always escapes this trapping point. Furthermore, when robot is trapped in complex obstacles (for example U-shape or long wall), the rotational vector field will help it find a new path to escape these obstacles. The direction of the rotational force can be clockwise or counter-clockwise (see Fig. 4). Hence, this rotational force is built as:

$$f_i^{or} = w_i^o c_i^o n_i^{or} \tag{12}$$

where the unit vector $n_i^{or}$ is given as

$$n_i^{or} = c_i^{or} \left( (y_i - y_o)/d_i^o, \quad -(x_i - x_o)/d_i^o \right)^T \tag{13}$$

where the scalar $c_i^{or}$ is used to define the direction for the rotational force: the rotational force is clock-wise if $c_i^{or} = 1$, and counter-clockwise if $c_i^{or} = -1$. Now, we consider the relationship between this unit vector and the vector $(p_i - p_o)$.

Let $\sigma$ be the angle between these vectors. Then, we have

$$\cos\sigma = \frac{c_i^{or}\left((x_i - x_o)(y_i - y_o) - (y_i - y_o)(x_i - x_o)\right)}{\left(d_i^o\right)^2} = 0. \tag{14}$$

Equation (16) shows that the unit vector $n_i^{or}$ is always perpendicular to vector $(p_i - p_o)$. The positive gain factor $w_i^o$ in (14) is used as a control element to drive robots to quickly escape obstacles. Therefore, this control element is designed such that the total force on robots always has the direction in the selected rotational direction. This control element is given as follows:

$$w_i^o = (1+c)\left(\|f_i^{top}\| + \lambda_o\right). \tag{15}$$

where $\lambda_o$ is a positive factor. The force $f_i^{top}$ is described as:

$$f_i^{top} = f_i^j + f_i^{op} \tag{16}$$

where $f_i^{op}$ is the repulsive force from the neighboring obstacles of robot $i$, and $f_i^j$ is the attractive force from the virtual node $j$ in the desired formation $f_i^j \approx u_i^j$, see Algorithm 1. The constant $c$, which depends on angle $\alpha$ between the sum vector $f_i^{top}$ and the unit vector $n_i^{or}$ (see Fig. 6), is described as follows:

$$c = \begin{cases} c_1, & \text{if } \alpha < \pi/2 \\ c_2, & \text{otherwise,} \end{cases} \tag{17}$$

where two constants $c_1$ and $c_2$ can be chosen as $-1 < c_1$, $0 < c_2$, and $c_1 < c_2$. Equation (15) guarantees that robot $i$ always moves in the direction of the rotational force $f_i^{or}$. Hence, it can easily escape the obstacles in order to continue with its swarm to track the moving target.

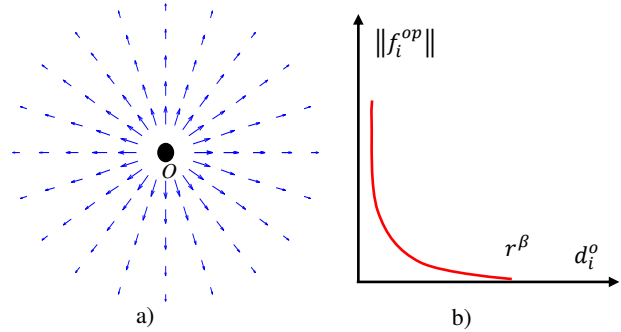

Fig. 3. The description of the repulsive force field $f_i^{op}$ surrounding the neighboring obstacle $o$ of the robot $i$ (a), and its amplitude $\|f_i^{op}\|$ (b).

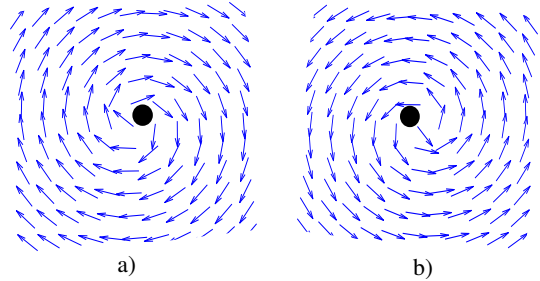

Fig. 4. The clockwise rotational force field (a) and the counter-clockwise rotational force field (b).

**Remark 3.** The protocol for collision avoidance for the agents of the system in Section 3.1 employs a repulsive force in (6). Here, the shapes of agents are simplified to be points or circles. In contrast, the obstacle avoidance in this section is realized using repulsive and rotational forces in (11) and (13) for complex obstacles (which can be large and nonconvex). This enables agents to escape the obstacles effectively.

### 3.5 Target tracking control algorithm

Firstly, a robot, which is closest to the target, is selected as the leader in order to generate the desired formation. Then, this leader leads its formation to track a moving target. When the leader encounters a risk, such as it is broken or trapped in obstacles, it must transfer its leadership to another, and becomes a free member robot in the swarm. The leader is selected as in Algorithm 2.

The target tracking controller, which is designed based on the relative position between the leader and the target, has to guarantee that the formation's motion is always driven towards the target. As introduced above, the

V-shape and circular shape formations are utilized to track and encircle a moving target. Hence, the tracking task is to make the distance between the leader and the target $d_l^t = \|p_l - p_t\|$ approaching the radius of the desired circular formation $r^\tau$ as fast as possible. This means that $\lim_{t\to\infty}(\|p_l(t) - p_t(t)\|) = r^\tau$ and $\lim_{t\to\infty}(\|v_l(t) - v_t(t)\|) = 0$. Based on the above analysis, the control law for the target tracking is proposed as follows:

$$u_l^t = f_l^t - k_{lv}^t(v_l - v_t) + \dot{v}_t \tag{18}$$

where $k_{lv}^t$ and $\dot{v}_t$ are the positive constant and the acceleration of the target, respectively, $(v_l - v_t)$ is the relative velocity vector between the leader and the target.

---

**Algorithm 2:** Leader selection

**Update data:** The actual position of robots $p_i$ ($i=1,..,N$, $i\neq l$), obstacle's information, the target's position $p_t$, the actual position of the leader ($p_\xi = p_l$).

**if** time $t=0$ (at initial time) **then**
    Compute the shortest distance from the robot $p_i$ to the target $p_t$ in order to determine the leader as
    $d_{imin1}^t = \min\{\|p_i - p_t\|, i=1,..,N\}$ ; $p_l = p_{imin1}$
**else**
    **if** the current leader meets obstacle or is broken **then**
        the leadership is transferred to a member that is free and has the closest distance to the target.
        $d_{imin2}^t = \min\{\|p_i - p_t\|, i=1,..,N, i \neq \xi, \text{free}\}$ ; $p_l = p_{imin2}$
    **else**
        Maintain the leadership of the current leader
        $p_l = p_\xi$
    **end**
**end**

---

The potential field $f_l^t$ from the target is used to drive the leader moving towards the target, and is given as:

$$f_l^t = \begin{cases} \left(\left(\dfrac{1}{d_l^t} - \dfrac{1}{r^\tau}\right)\dfrac{k_{l2}^t}{(d_l^t)^2} - \dfrac{k_{l3}^t(d_l^t - r^\tau)}{(r^t - r^\tau)}\right) n_l^t, & \text{if } d_l^t < r^t \\ -k_{l1}^t n_l^t, & \text{otherwise,} \end{cases} \tag{19}$$

where $k_{l2}^t$, $k_{l3}^t$ are positive constants, $r^t$ and $r^\tau$ are the target approaching radius, and the desired radius of the circular formation $r_{min}^\tau < r^\tau < r^t$, respectively. Here $k_{l1}^t = k_{l1d}^t + \varepsilon_2 \|v_t\|$ where $k_{l1d}^t$, $\varepsilon_2$ are positive constants. The unit vector along the line connection from the target to the leader is computed as $n_l^t = (p_l - p_t)/\|p_l - p_t\|$. In equation (19), the constant attractive force $f_{2l}^t = -k_{l1}^t n_l^t$ is used to track the target when $d_l^t > r^t$. On the other hand, the attractive/repulsive force field surrounding the equilibrium position, at which $\|p_l - p_t\| = r^\tau$ and

$(v_l - v_t) = 0$, is employed to encircle the target when $d_l^t \leq r^t$. Hence, using this combined vector field, the leader can easily approach the target at the equilibrium position.

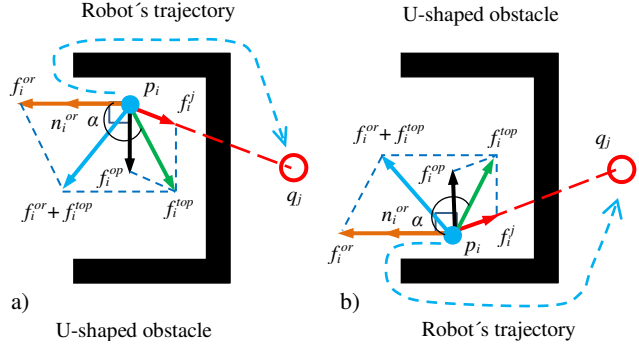

Fig. 5. Description of the obstacle escape of a robot *i* while reaching a virtual node *j*: clockwise (a) and counter-clockwise (b).

**Theorem 3.** *Consider the leader l described by model (1) and governed by control law (18) when $d_l^t \leq r^t$. If the velocity of the target is smaller than the maximum velocity of the leader, then the system (1) will converge to the equilibrium state, at which $v_l = v_t$ and $(p_l - p_t) = r^\tau (p_l - p_t)/\|p_l - p_t\|$.*

Proof: See the proof of the theorem in Appendix.

**Remark 4.** The motion of the formation depends on the relative position between the leader and the target. At the initial time, a robot, which is closest to the target, is chosen as a leader to lead its formation towards the target. During its movement, if the current leader encounters any risk (for instance it is broken from the formation or hindered by the environment), then a new leader is nominated. Algorithm 2 guarantees that this new leader will reorganize the formation and continue to lead the new formation to track the target.

**Remark 5.** Obstacles in the environment can be sensed by several methods. Onboard sensors of each agent can detect them. Hybrid approaches can employ a group of aerial drones to capture the images of obstacles and agents and transfer data to the agents [42, 43].

## 4 CASE STUDY

The proposed algorithms as discussed in the previous section can allow the robots to follow predefined formations such as V-shape, circular or line shape. Due to their similarity, we just present two cases of V-shape and circular shape. Let $p_l = (x_l, y_l)^T$ and $p_t = (x_t, y_t)^T$ be the positions of the leader and the target, respectively. The relative position vector between the leader and the target is $(p_l - p_l) = (x_l - x_l, y_l - y_l)^T$, and the distance between them is determined as $d_l^t = \sqrt{(x_l - x_t)^2 + (y_l - y_t)^2}$.

### 4.1 V-shape desired formation

Firstly, in order to build the V-shape desired formation, we design the right side of this V-shape based on the desired formation angle $\varphi_d$ and the relative position between the leader and the target. As depicted in Fig. 6, the coordinates of the base node $q_\mu$ on the coordinate system $x'y'$ ($q'_\mu = (x'_\mu, y'_\mu)^T$) are determined as follows:

$$\begin{pmatrix} x'_\mu \\ y'_\mu \end{pmatrix} = \|q_\mu - p_l\| \begin{pmatrix} \cos \delta_d \\ \sin \delta_d \end{pmatrix} \tag{20}$$

where the angle $\delta_d = \angle(q_\mu - p_l), (p_t - p_l)$ is equal to $\varphi_d/2$ and $\varphi_d > 0$ is the desired formation angle, see Fig. 6. By rotating and translating equation (20) according to coordinate systems $x''y''$ and $xy$, see [37], we obtain the position of the desired node $q_\mu$ on the coordinate system $xy$ as follows:

$$q_\mu = p_l + R q'_\mu \tag{21}$$

From base node $q_\mu$ and the position of the leader, we determine a unit vector along the line connecting from $q_\mu$ to $p_l$ as $n_{\mu l} = (p_l - q_\mu)/\|p_l - q_\mu\|$. The rotational matrix $R$, which depends on the rotational angle theta $\theta$, is determined as

$$R = \begin{cases} \begin{pmatrix} \cos \theta & -\sin \theta \\ \sin \theta & \cos \theta \end{pmatrix} & \text{if } \theta \text{ rotates clockwise,} \\ \begin{pmatrix} \cos \theta & \sin \theta \\ -\sin \theta & \cos \theta \end{pmatrix} & \text{otherwise.} \end{cases} \tag{22}$$

Now, a virtual node $j$ ($d^l_j = \|p_l - q_j\| = jd$; $q_j = (x_j, y_j)^T$; $v_j = (v_{jx}, v_{jy})^T$; $j = 1,2,..,N^*$; $N^* \in R$) is determined by the unit vector $n_{\mu l}$ as

$$(q_j - p_l) = jd n_{\mu l}. \tag{23}$$

Substituting $n_{\mu l} = (p_l - p_\mu)/\|p_l - p_\mu\|$ into equation (21), we obtain:

$$q_j = (1+j)p_l - j q_\mu. \tag{24}$$

Equation (24) can be rewritten as

$$\begin{pmatrix} x_j \\ y_j \end{pmatrix} = \begin{pmatrix} (1+j)x_l \\ (1+j)y_l \end{pmatrix} - \begin{pmatrix} j x_\mu \\ j y_\mu \end{pmatrix}. \tag{25}$$

Equation (25) shows that when $j$ changes from $j=1$ to $j=N^*$ we obtain the formation of the $N^*$ virtual nodes. These nodes lie on a line connecting $p_l$ and $q_\mu$, and are equally spaced (the right side of the desired V-shape formation), see Fig. 6.

Similarly, the virtual nodes $j$ on the left side of the V-shape is designed as

$$(q_j - p_l) = jd n_{l\eta}. \tag{26}$$

where $n_{l\eta} = (q_\eta - p_l)/\|q_\eta - p_l\|$ is a unit vector. Substituting this unit vector into (26), we obtain

$$q_j = (1-j)p_l + j q_\eta. \tag{27}$$

In Equation (27), $q_\eta$ is the position of the base node on the line which deviates from the line through the leader and the target with an angle of $(\pi - \delta_d)$, see Fig. 6. Similar to equation (28), this base node is determined as

$$q_\eta = p_l + R q'_\eta. \tag{28}$$

Using (28), we obtain the formation of the virtual nodes $j$ ($j = 1,2,..,N^*$). They lie on the line through $p_l$ and $q_\eta$, and are equally spaced (the left side of the V-shape), see Fig. 6.

Finally, the law to generate the desired V-shape formation of the virtual nodes $j$ ($j=1,2..,N$) is proposed as:

$$q_j = \begin{cases} (1+\xi_1)p_l - \xi_1 q_\mu, & \text{if } j \leq N/2 + 1 \\ (1-\xi_2)p_l + \xi_2 q_\eta, & \text{otherwise,} \end{cases} \quad (29)$$

where $\xi_1 = j - 1$ and $\xi_2 = j - 1 - floor(N/2)$ are the positive factors. Using equation (29), the virtual nodes $j$ ($j=1,2,...,N$) will be evenly distributed to both sides of the leader. Hence, we obtain a V-shape formation due to these desired virtual nodes and a constant formation angle $\varphi_d = 2\delta_d$ as depicted in Fig.8a. In some cases, such as under the influence of an environmental factor (noises, wind, obstacle avoidance, etc.,), the formation angle needs to be changed to adapt to the effect of this environment. Thus, the formation angle is chosen as

$$\varphi(t) = \varphi_d + \varepsilon_3 \phi(t), \quad (30)$$

where $\varepsilon_3$ is a positive constant, and $\phi(t)$ is used as a sensing function that decides the formation angle $\varphi(t)$. However, this formation angle $\varphi(t)$ must guarantee that there are no collisions among the members in the formation. In other words, it depends on the repulsive radius of each robot. Hence, the smallest formation angle is computed as $\varphi_{dmin} = \arccos(1 - r_r^2 / 2d^2)$.

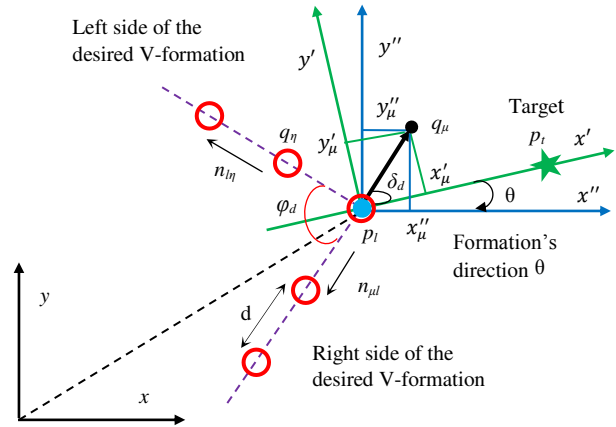

Fig. 6. The description of the method to build the V-shape desired formation.

4.2 Circular desired formation

As presented above, the circular formation is used to encircle the moving target when the distance between the leader and the target is shorter than the target approaching radius $d_l^t = \|p_l - p_t\| \leq r^t$. Hence, this desired formation is designed based on the relative position between the target and the leader as shown in Fig. 8.

The position of the virtual node $j$ ($j=1,2..,N$) on the circle, whose central point is at the target's position $p_t$ and whose radius is $d_l^t = \|p_l - p_t\|$, is computed as

$$q_j = p_t + Rq_j', \quad j = 1,2..,N. \quad (31)$$

The position of the virtual node $j$ on the coordinate system $x'y'$ ($q'_j = (x'_j, y'_j)^T$) is computed as

$$\begin{pmatrix} x'_j \\ y'_j \end{pmatrix} = d^t_l \begin{pmatrix} \cos(2j\pi/N) \\ \sin(2j\pi/N) \end{pmatrix}, \quad j = 1,2..,N. \tag{32}$$

Let the virtual node owned by the leader be the first position in the circular desired formation. Substituting equation (32) into (31), we have the circular formation of the virtual nodes $j$ as follows:

$$\begin{pmatrix} x_j \\ y_j \end{pmatrix} = \begin{pmatrix} x_t \\ y_t \end{pmatrix} + R d^t_l \begin{pmatrix} \cos(2\zeta\pi/N) \\ \sin(2\zeta\pi/N) \end{pmatrix}, \quad j = 1,2..N; \zeta = j-1. \tag{33}$$

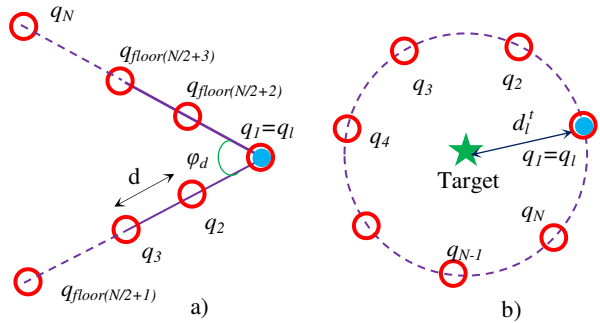

Fig. 7. The description of the distributed virtual nodes $j$ in the V-shape desired formation (a) and in the circular shape desired formation (b).

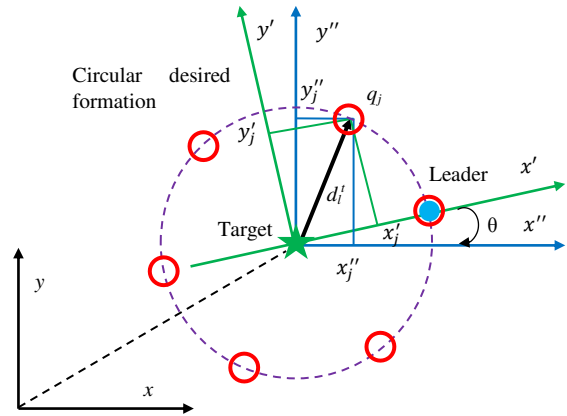

Fig. 8. The description of the method to build the circular desired formation.

Equation (33) shows that the distributed virtual nodes $j$ $(j=1,2,..,N)$ on the circular desired formation are equidistant, and the distance from them to the target is $d_l^t = \|p_l - p_t\|$, see Fig. 7b. The desired radius is chosen such as the distances among agents guarantee collision avoidance when the circle formation is achieved.

## 5 SIMULATION RESULTS

In this section, we use the V-shape and circular desired formations, which are built in Section 4, to test the proposed control algorithms. We assume that the initial velocities of the robots and the target are zero. The initial positions of the robots are random. Each robot can sense the position of other robots within its sensing range as well as the positions of the target and obstacles. The target moves in a sine wave trajectory, which is defined as $p_t = (0.9t + 640, \; 160\sin(0.01t) + 250)^T$. The general parameters of the simulations are listed in Table II.

TABLE II: PARAMETER VALUES

| Parameter | Definition | Value |
|---|---|---|
| $N$ | Number of robots | 9 |
| $\varphi_d$ | Desired formation angle | $2\pi/3$ rad |
| $r^\tau$ | Desired radius of circular formation | 60 m |
| $r^t$ | Target approach radius | 100 m |
| $r_r$ | Collision radius around each robot | 45 m |
| $d$ | Desired distance between robots in the V-shape | 60 m |
| $\lambda, \lambda^*, \lambda_a$ | Positive constants | 15, 9, 5 |
| $\varepsilon_1, \varepsilon_2, \varepsilon_3$ | Positive constants | 1, 0.5, 0.7 |
| $k_{i1}^k, k_{i2}^k$ | Constants for fast repulsion | 80, 12 |
| $k_{i1}^o, k_{i2}^o$ | Constants for fast obstacle avoidance | 90, 15 |
| $k_{l1d}^t, k_{l2}^t, k_{l3}^t$ | Constants for approaching to target | 1, 0.6, 4 |
| $k_{ip1}^j, k_{ip2}^j, k_{ip3}^j$ | Positive constants | 3, 4, 9 |
| $k_{iv}^j, k_{iv}^t, k_{iv}^k$ | Damping constants | 1, 1.2, 1.5 |

### 5.1 The connection of the swarm

For this simulation, the formation angle is selected $\varphi(t) = 2\pi/3$. The proposed algorithms were used to generate the desired formations (V-shape and circular shape), and control the robots to move towards the virtual nodes in the desired formation (Algorithm1).

The results in Fig. 9 show that the desired formations were easily created. Robots, which have random initial positions, quickly achieved the desired positions in these desired formations while tracking a moving target without collisions. The position permutations among the members in the formation appeared, but they did not influence the structure of the formation when tracking the target. As shown in Fig. 9, at the initial time, one robot was chosen as the leader to drive its formation towards the target in a V-shape formation. At time $t=70s$, the V-shape formation was constructed, and it was kept until the square robot detected the obstacle $O_1$. At time $t=160s$, while avoiding the obstacle $O_1$, the virtual node, which was owned by the square robot, became a free node. Then, this virtual node attracted the triangular robot to become the active node at time $t=200s$. After escaping the obstacle, the square robot quickly approached the remaining free node of the desired formation as shown in Fig. 9. Similarly, the rhombus robot was permuted with other robots in the formation while avoiding the obstacle $O_2$. At time $t=250s$, the V-shape formation changed to the circular formation to encircle the target. In this situation, the member robots became the free robots, and then they approached the desired circular formation to become. Fig. 10 shows that, this circular formation was kept around the target at the desired radius $r^\tau$ at time $t=320s$. In other words, the leader's position is stable at the equilibrium point, at which $\|p_l - p_t\| = r^\tau$.

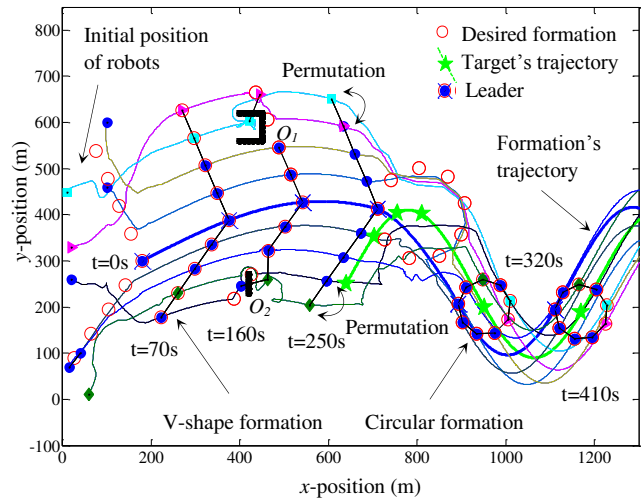

Fig. 9. Evolution of a swarm following the desired formations under the influence of the dynamic environment while tracking a moving target.

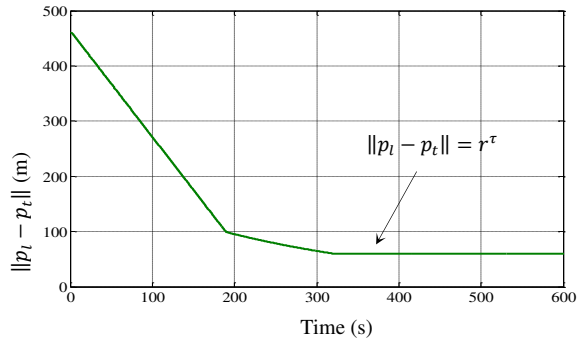

Fig. 10. Distance between the leader and the target in case the leader is not hindered while tracking a moving target.

Next we consider the connection of a swarm when the leader was trapped in the complex obstacle (for example U-shape obstacle, see Fig. 11). In this situation, the actual leader had to transfer its leadership to other members in the swarm, and managed to escape this obstacle. Fig. 11 shows that, at time $t=0s$, the square robot was chosen as the leader, and its leadership was kept until it was trapped in the U-shape obstacle at time *t=200s*. While avoiding obstacle, the square leader transferred its leadership to the triangular robot, which was not faced with any obstacles and closest to the target. Then, this square leader became a free robot. It automatically found a way (lilac way) to escape this U-shape obstacle in order to continue following its formation.

After receiving the leadership, the triangular robot reorganized a new formation, and continued to lead this formation to track the target. The distance between the new leader and the target shrunk until it achieved the active radius of the circular desired formation $\|p_l - p_t\| = r^\tau$. Then, this distance was maintained to encircle the moving target, see Fig. 12. Moreover, Fig. 11 shows that the position permutation between the square leader and the triangular leader did not influence the structure of the formation.

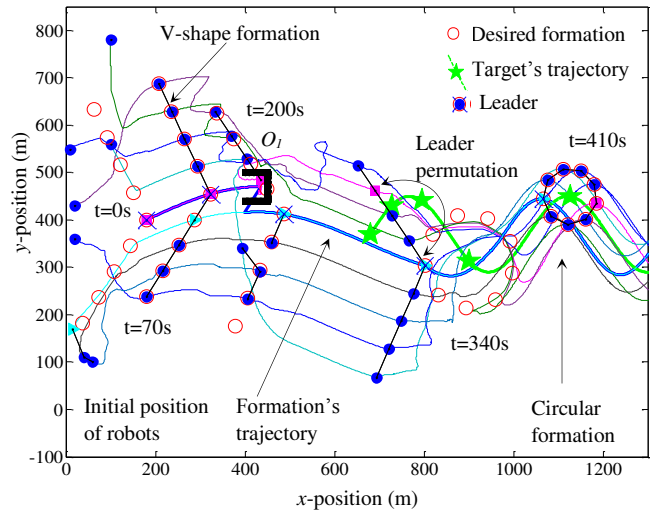

Fig. 11. Path planning for a swarm following the desired formations while tracking a moving target with the leader permutation.

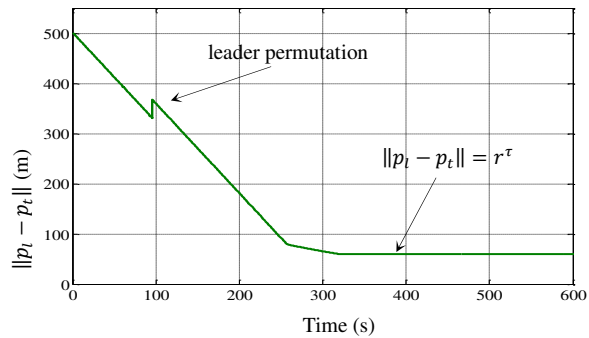

Fig. 12. Distance between the leader while tracking a moving target in case the leader is permuted.

## 5.2 The robustness of the swarm in noisy environment

In this subsection, we examine the robustness of the formation under the influences of noise and the change of the formation angle. The noises used in this simulation are the Gaussian function with zero mean, variance of 1 and standard deviation of 1, see Fig. 13. The formation angle $\varphi(t)$ is used for simulations as $\varphi(t) = 2\pi/3 + 1.6\sin(0.2t)$, see Fig. 14.

Fig. 16 shows that robot $i$ was always close to active node $j$ in the desired formation, and its formation was maintained following the desired formations (V-shape and circular formation) under the influence of the noisy environment and the changes of the formation angle $\varphi(t)/2$. The position error between each robot $i$ and active node $j$, at which this robot $i$ was occupying, is small, see Fig. 14 and Fig. 15. Fig. 16 also reveals that from the random initial positions, the free robots quickly found their desired position in the desired V-formation. Then, they tracked a moving target in a stable V-formation. At time $t=70s$, under the influence of the sudden change of the formation angle from $\varphi=2\pi/3$ to $\varphi=(\pi-0.6)$, the formation was broken, but it was quickly redesigned to continue to track the moving target. In contrast, when the formation angle $\varphi(t)$ changed slowly, the formation of robots was always maintained following the desired V-formation with the small position errors, see Fig. 15, Fig. 16. Moreover, the simulation results show that the noise had influences on the position errors of the robot formation, but they only caused small formation changes as shown in Fig. 15.

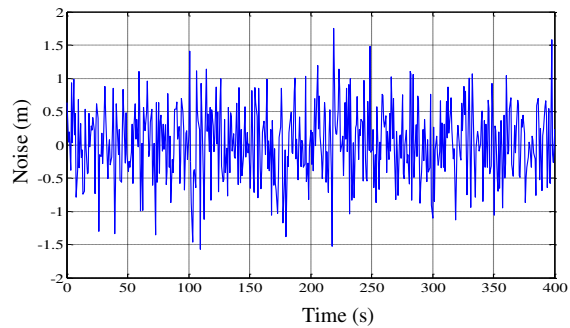

Fig. 13. Noise effects on the system.

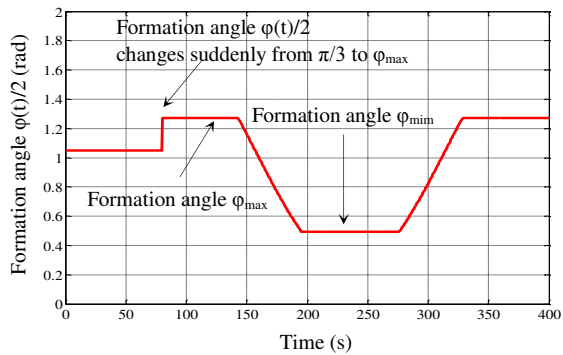

Fig. 14. Formation angle $\varphi(t)/2$ while tracking a moving target.

## 6 CONCLUSION

This paper presents a novel approach to formation control of autonomous robots following desired formations to track a moving target in a dynamic and noisy environment. The robot team can form predefined formations such as V-shape or circular shape while tracking a moving target. The stability and convergence analysis of the proposed formation control is given. The rotational force field combining with the repulsive force can drive the robot to quickly escape the obstacles, and more importantly is to avoid the local minimum problems when the sum of the attractive and repulsive forces of the potential field is equal to zero in the case of concave obstacle shapes.

The development and application of this proposed approach for formation control of the flight robots in 3D space, such as formation of the unmanned aerial vehicles will be our future research.

ACKNOWLEDGMENTS

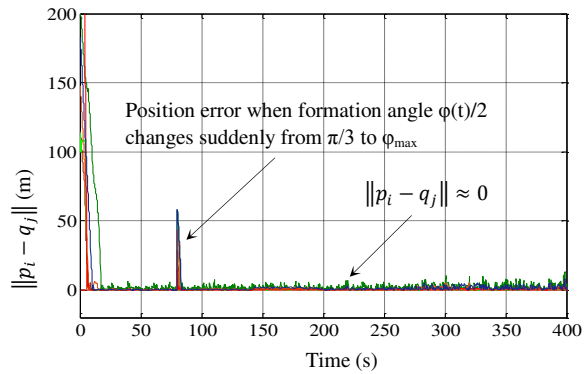

Fig. 15. Position errors $\|p_i - q_j\|$ under the effect of the noise.

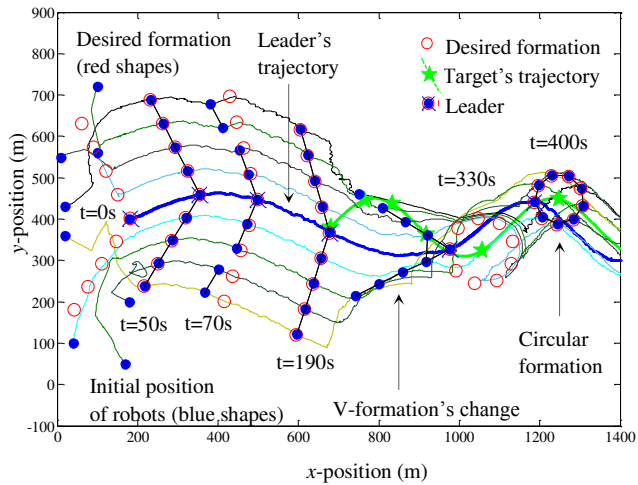

Fig. 16. The influences of noises and the different formation angles on the swarm's trajectory while tracking a moving target.

APPENDICES

In this appendix, we analyze and prove the stability of the proposed formation control algorithms.

A. Proof of Theorem 2

In order to analyze the stability of the robot $i$ at the active node $j$ when the node $j$-$1$ is also active, we rewrite the control law $u_i^j$ as follow:

$$u_i^j = f_{2i}^j - k_{iv}^j(v_i - v_j) + \dot{v}_j. \tag{34}$$

Consider the vector field $f_{2i}^j = -k_{ip1}^j(p_i - q_j) = (P_i, Q_i, Z_i)^T$, here $P_i = -k_{ip1}^j(x_i - x_j)$, $Q_i = -k_{ip1}^j(y_i - y_j)$ and $Z_i = 0$. Ac-cording to [38], we obtain:

$$rot(f_{2i}^j) = \left(\frac{\partial Z_i}{\partial y_i} - \frac{\partial Q_i}{\partial z_i}, \ \frac{\partial P_i}{\partial z_i} - \frac{\partial Z_i}{\partial x_i}, \ \frac{\partial Q_i}{\partial x_i} - \frac{\partial P_i}{\partial y_i}\right)^T = \mathbf{0}. \tag{35}$$

Equation (42) shows that the vector field $f_{2i}^j$ is irrotational. Consider the scale function as follows:

$$V_i^j = \frac{1}{2}k_{ip1}^j(p_i - q_j)^T(p_i - q_j). \tag{36}$$

Taking the negative gradient of the function $V_i^j$, we obtain:

$$-\nabla V_i^j = -\nabla\left(\frac{1}{2}k_{ip1}^j(p_i - q_j)^T(p_i - q_j)\right) = -\nabla\left(\frac{1}{2}k_{ip1}^j\left((x_i - x_j)^2 + (y_i - y_j)^2\right)\right)$$

$$= \begin{pmatrix} -\frac{1}{2}k_{ip1}^j\frac{\partial}{\partial x_i}\left((x_i - x_j)^2 + (y_i - y_j)^2\right) \\ -\frac{1}{2}k_{ip1}^j\frac{\partial}{\partial y_i}\left((x_i - x_j)^2 + (y_i - y_j)^2\right) \end{pmatrix}^T = \begin{pmatrix} -k_{ip1}^j(x_i - x_j) \\ -k_{ip1}^j(y_i - y_j) \end{pmatrix}^T = -k_{ip1}^j(p_i - q_j) = f_{2i}^j. \tag{37}$$

So, (35) and (37) show that the vector field $f_{2i}^j$ is a potential field, and its potential function is $V_i^j$.

Let $x_1 = p_i - q_j$, $x_2 = v_i - v_j$ be the relative position and velocity of the robot $i$ and node $j$. We have the error dynamic model of the system as follows:

$$\begin{aligned} \dot{x}_1 &= x_2 \\ \dot{x}_2 &= u_i^j - \dot{v}_j \quad i, j = 1, 2.., N. \end{aligned} \tag{38}$$

Substituting $u_i^j$ in (41) into (45), we obtain the error dynamic model of the system as:

$$\dot{x} = Ax \tag{39}$$

where $x = (x_1, x_2)^T$ and

$$A = \begin{bmatrix} 0_2 & I_2 \\ -k_{ip1}^j I_2 & -k_{iv}^j I_2 \end{bmatrix} \tag{40}$$

where $0_2$ is the zero matrix of 2 and $I_2$ is the identity matrix of 2. It can easily be shown that the eigenvalues of matrix $A$ are the roots of the polynomial $s^2 + k_{ip1}^j s + k_{lv}^j$, which have negative real parts since $k_{ip1}^j$ and $k_{lv}^j$ are positive. Hence the system (39) is asymptotically stable [47].

### B. Proof of Theorem 3

Firstly, we consider the vector field $f_{ll}^t = K_l n_l^t$ in (21), here $K_l = (k_{l2}^t/d_l^t - k_{l2}^t/r^\tau)/(d_l^t)^2 - k_{l1}^t(d_l^t - r^\tau)/(r^t - r^\tau)$. Let $P_l = K_l(x_l - x_t)/\|p_l - p_t\|$, $Q_l = K_l(y_l - y_t)/\|p_l - p_t\|$, and $Z_l = 0$, we have:

$$rot(f_{ll}^t) = \left( \frac{\partial Z_l}{\partial y_l} - \frac{\partial Q_l}{\partial z_l},\ \frac{\partial P_l}{\partial z_l} - \frac{\partial Z_l}{\partial x_l},\ \frac{\partial Q_l}{\partial x_l} - \frac{\partial P_l}{\partial y_l} \right)^T = \mathbf{0}. \tag{41}$$

Equation (41) shows that the vector field $f_{ll}^t$ is not rotational.

Consider the scale function as follow:

$$V_l^t = \frac{1}{2}\left( k_{l2}^t \left( \frac{1}{d_l^t} - \frac{1}{r^\tau} \right)^2 + \frac{k_{l1}^t(d_l^t - r^\tau)^2}{(r^t - r^\tau)} \right). \tag{42}$$

Taking the negative gradient of $V_i^t$, we obtain:

$$f_{ll}^t = -\nabla V_l^t = -\nabla \left( \frac{k_{l2}^t}{2}\left(\frac{1}{d_l^t} - \frac{1}{r^\tau}\right)^2 + \frac{k_{l1}^t(d_l^t - r^\tau)^2}{2(r^t - r^\tau)} \right) = -\left( \frac{k_{l2}^t}{d_l^t} - \frac{k_{l2}^t}{r^\tau} \right) \nabla\left( \frac{1}{d_l^t} - \frac{1}{r^\tau} \right) - \frac{k_{l1}^t(d_l^t - r^\tau)}{(r^t - r^\tau)} \nabla(d_l^t - r^\tau)$$

$$= \left( \frac{k_{l2}^t}{d_l^t} - \frac{k_{l2}^t}{r^\tau} \right)\frac{1}{(d_l^t)^2}\nabla d_l^t - \frac{k_{l1}^t(d_l^t - r^\tau)}{(r^t - r^\tau)}\nabla d_l^t = \left( \left( \frac{1}{d_l^t} - \frac{1}{r^\tau} \right)\frac{k_{l2}^t}{(d_l^t)^2} - \frac{k_{l1}^t(d_l^t - r^\tau)}{(r^t - r^\tau)} \right)\nabla d_l^t. \tag{43}$$

Similar to equation (37), we obtain a gradient field $\nabla d_l^t = (p_l - p_t)/\|p_l - p_t\|$. So, (41) and (43) show that the vector field $f_{ll}^t$ is also a potential field, and its potential function is $V_l^t$.

In order to analyze the stability of the leader at the equilibrium position, at which $\|p_l - p_t\| = r^\tau$ and $(v_l - v_t) = 0$, let $\breve{x}_1$ be the vector $(p_l - p_t)$, and $\breve{x}_2$ be the relative velocity between the leader and the target. Then, the error dynamic of the system is described as

$$\dot{\breve{x}}_1 = \breve{x}_2$$
$$\dot{\breve{x}}_2 = \dot{v}_l - \dot{v}_t. \tag{44}$$

Substituting (18) into (44), we obtain:

$$\dot{\breve{x}}_1 = \breve{x}_2$$
$$\dot{\breve{x}}_2 = -\nabla V_l^t - k_{lv}^t \breve{x}_2. \tag{45}$$

To analyze the stability of the system (45), we choose the following Lyapunov function candidate

$$V_\tau = V_l^t + \frac{1}{2}\breve{x}_2^T \breve{x}_2. \tag{46}$$

Taking the time derivative of equation (46) along the trajectory of the system (45), we obtain:

$$\dot{V}_\tau(t) = \nabla V_i^t \dot{\breve{x}}_1 + \breve{x}_2^T \dot{\breve{x}}_2 = \breve{x}_2(\nabla V_i^t + \dot{\breve{x}}_2) = -k_{lv}^t \breve{x}_2^T \breve{x}_2 \leq 0. \tag{47}$$

According to LaSalle's Theorem, $\breve{x}_2 \to 0$, [47]. It can easily be shown that $\dot{\breve{x}}_2$ is uniformly continuous. Hence, according to Barbalat's Lemma, $\dot{\breve{x}}_2 \to 0$, [48]. This implies from (45) that $\nabla V_l^t \to 0$ or $f_l^t \to 0$. Hence, $d_t^l \to r^\tau$ due to the description of the potential force function in (19). So, the system (45) is stable at the equilibrium position when using the control law (18). In other words, using the controller (18), the leader will converge to the equilibrium position, at which the distance between leader and the target is equal to the radius of the desired circular formation.